\documentclass[conference,a4paper]{IEEEtran} 
\IEEEoverridecommandlockouts
\usepackage{cite}
\usepackage{amsmath,amssymb,amsfonts}
\usepackage{algorithmic}
\usepackage{graphicx}
\usepackage{subfigure}
\usepackage{textcomp}
\usepackage{hyperref}
\usepackage{xcolor}
\pdfoutput=1
\def\BibTeX{{\rm B\kern-.05em{\sc i\kern-.025em b}\kern-.08em
    T\kern-.1667em\lower.7ex\hbox{E}\kern-.125emX}}
\begin{document}

\title{Scalable Label-efficient Footpath Network Generation Using Remote Sensing Data and Self-supervised Learning}

\author{
    \IEEEauthorblockN{Xinye Wanyan$^{a}$, Sachith Seneviratne$^{a,b}$, Kerry Nice$^{a}$, Jason Thompson$^{a}$, Marcus White$^{c}$, 
    \\ Nano Langenheim$^{a}$, Mark Stevenson$^{a,b}$}
    \IEEEauthorblockA{$^a$ Transport, Health, and Urban Design Research Lab, Melbourne School of Design, University of Melbourne, Australia}
    \IEEEauthorblockA{$^b$ Faculty of Engineering and Information Technology, University of Melbourne, Australia}
    \IEEEauthorblockA{$^c$ Centre for Design Innovation, Swinburne University of Technology, Australia}
    \IEEEauthorblockA{\{x.wanyan, sachith.seneviratne, kerry.nice, jason.thompson, Nano Langenheim, mark.stevenson\}@unimelb.edu.au, \\
    marcuswhite@swin.edu.au}
}

\maketitle

\begin{abstract}
   Footpath mapping, modeling, and analysis can provide important geospatial insights to many fields of study, including transport, health, environment and urban planning. The availability of robust Geographic Information System (GIS) layers can benefit the management of infrastructure inventories, especially at local government level with urban planners responsible for the deployment and maintenance of such infrastructure.
   However, many cities still lack real-time information on the location, connectivity, and width of footpaths, and/or employ costly and manual survey means to gather this information.
   This work designs and implements an automatic pipeline for generating footpath networks based on remote sensing images using machine learning models.
   The annotation of segmentation tasks, especially labeling remote sensing images with specialized requirements, is very expensive, so we aim to introduce a pipeline requiring less labeled data. 
   Considering supervised methods require large amounts of training data, we use a self-supervised method for feature representation learning to reduce annotation requirements.
   Then the pre-trained model is used as the encoder of the U-Net for footpath segmentation.
   Based on the generated masks, the footpath polygons are extracted and converted to footpath networks which can be loaded and visualized by geographic information systems conveniently. Validation results indicate considerable consistency when compared to manually collected GIS layers.
   The footpath network generation pipeline proposed in this work is low-cost and extensible, and it can be applied where remote sensing images are available.
   Github: \href{https://github.com/WennyXY/FootpathSeg}{https://github.com/WennyXY/FootpathSeg}.
\end{abstract}


\section{Introduction}
\begin{figure}
  \centering
  \subfigure{
    \centering
    \includegraphics[width=0.65\linewidth]{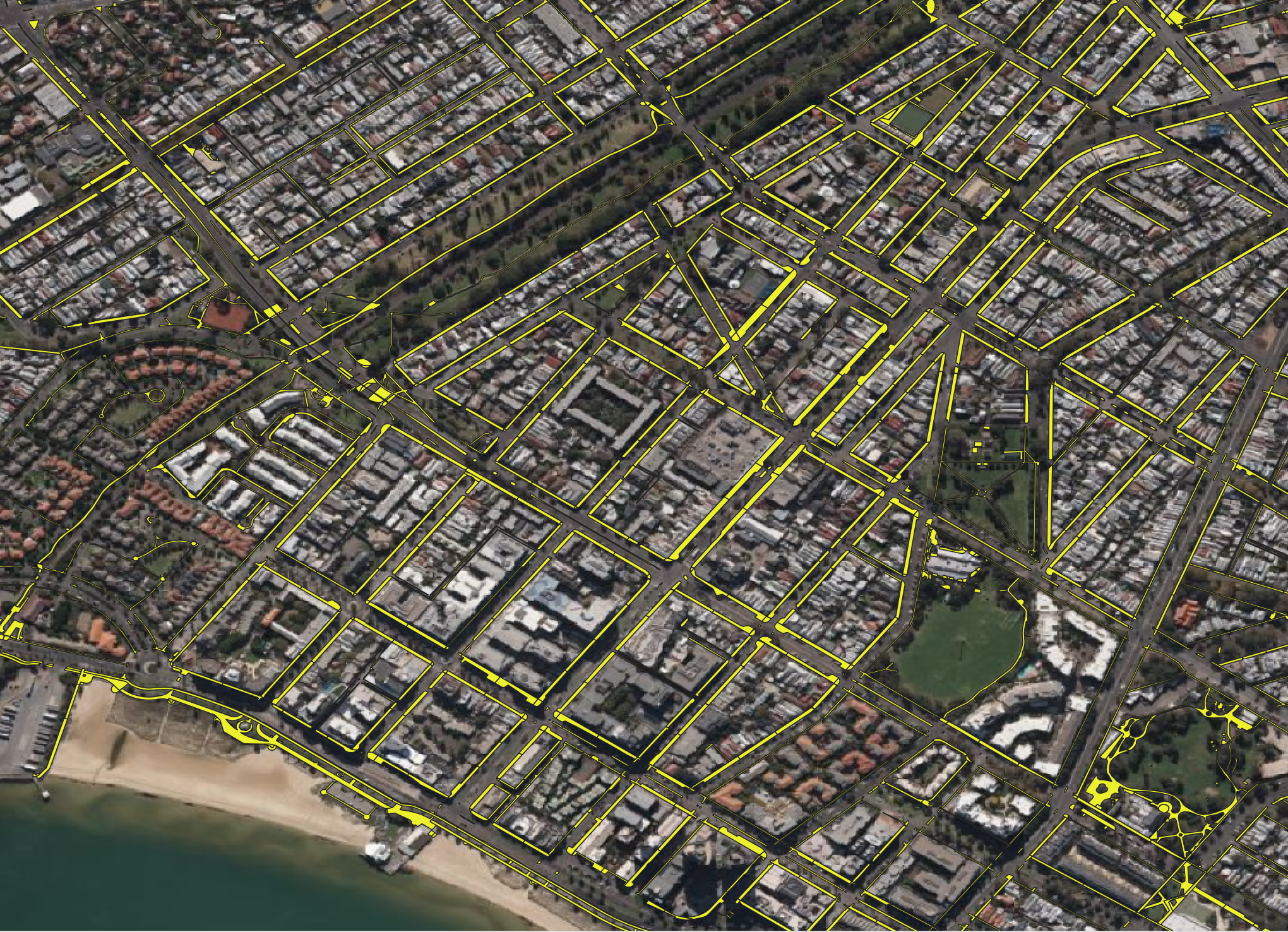}
  }
  \subfigure{
    \centering
    \includegraphics[width=0.65\linewidth]{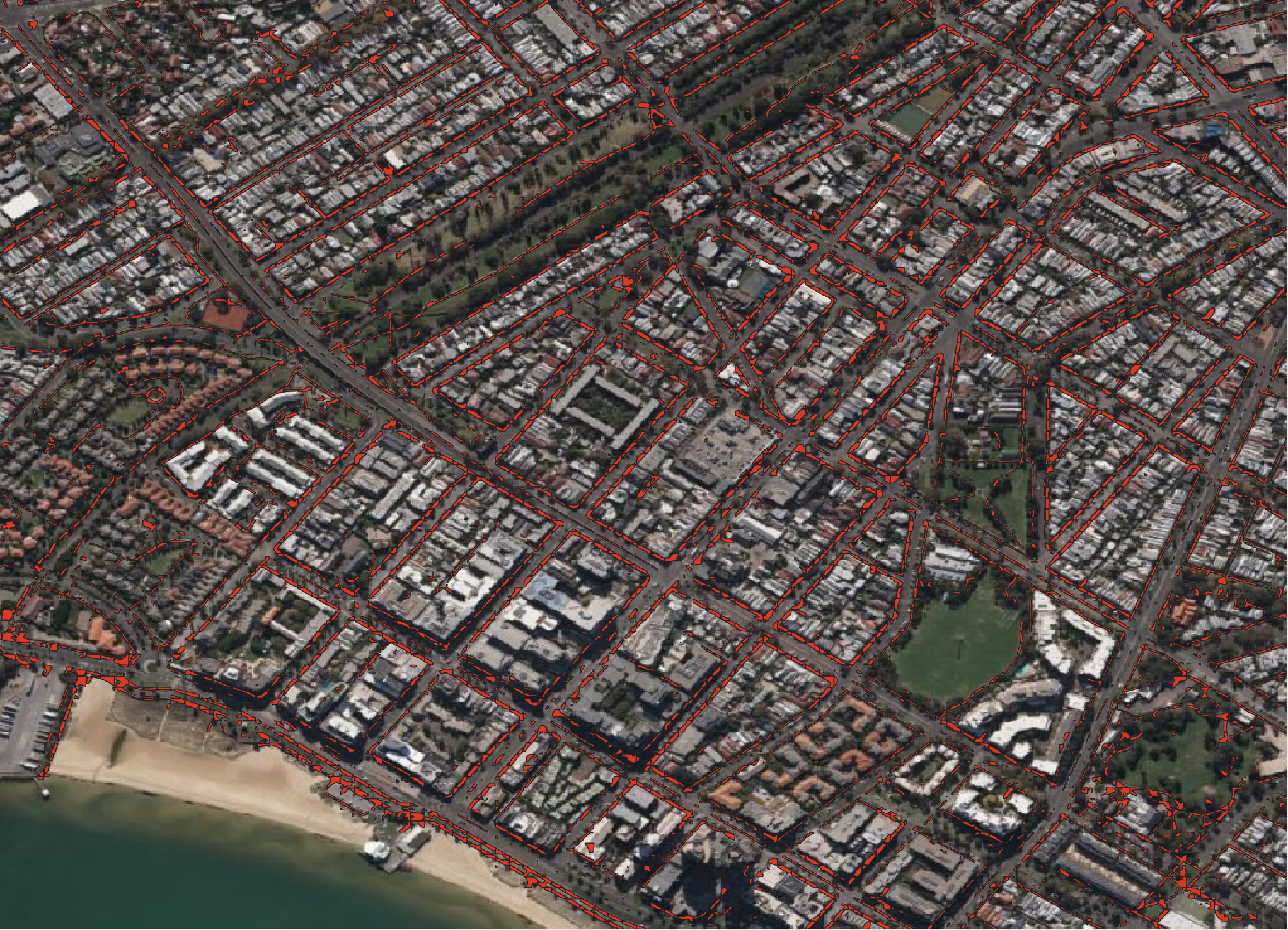}
  }
  \caption{
  An example of the ground truth footpath network in an area of Melbourne.
  The figure on the top is the ground truth footpath network and the figure on the bottom is generated by our pipeline automatically.
  The background map is provided by Bing Aerial.
  }
  \label{fig:intro}
\end{figure}
\begin{figure*}
  \centering
  \includegraphics[width=0.9\linewidth]{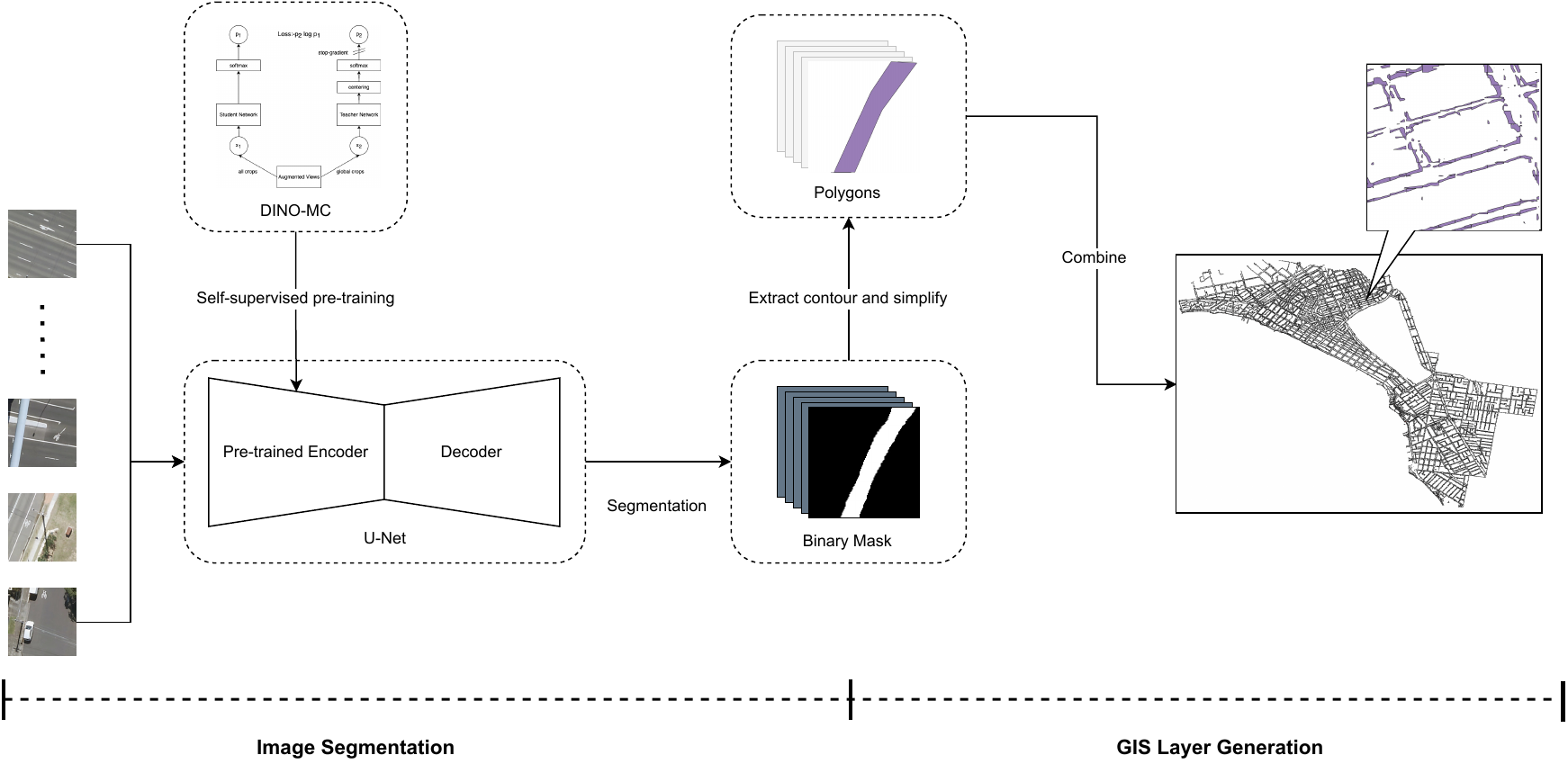}
  \caption{
  The pipeline of generating geographic footpath network based on the remote sensing images.
  The input of the pipeline is a series of remote sensing images of a specific area and the output is a generated footpath network about this area.
  First, we use an unlabeled remote sensing dataset to pre-train a feature extractor in a self-supervised learning method.
  Then, we fine-tune the U-Net to do the downstream segmentation task on a labeled footpath remote sensing dataset.
  The encoder of the U-Net is a pre-trained backbone network of the self-supervised model which is mainly used to generate the representations for the input, while the decoder is a custom convolutional neural network that takes the extracted features from the encoder and reconstructs the segmentation masks with the same dimensions as the input images.
  }
  \label{fig:pipeline}
\end{figure*}
Walking as a mode of transportation provides many health and economic benefits and is crucial for accessible transport systems \cite{frank2006many, lo2009walkability, hosseini2023mapping, white2022people}.
While city planners and local government agencies make significant investments in walking infrastructure, there is a lack of robust, real-time geolocated data capturing network detail. 
Such data is paramount for analysis such as walkability, safety, efficiency and maintenance of walking-oriented infrastructure.

Footpath networks require detailed and regular assessment, monitoring, and updating \cite{gm2021urban} to identify network flaws and potential risks to safety and accessibility ahead of time \cite{fotios2018illuminance}.
There are several existing efforts to detect and segment footpaths using Bird’s-Eye-View (BEV) or images captured by mobile phone applications \cite{gm2021urban, gosala2023skyeye}.
However, the data availability of such crowd-sourced methods is limited because captured images depend on the paths more frequently used by application users and are difficult to extend to more regions.
Instead, the development of remote sensing technology offers an opportunity as there is a large amount of open-access aerial imagery available to researchers.
Therefore, the effective use of remote sensing images to generate map information has a high potential for exploration.
Combined with recent advances in computer vision models that have demonstrated their effectiveness in object detection, there is an opportunity to apply vision models to solve existing footpath identification and monitoring issues.
However, annotation is very expensive, time-consuming, and error-prone, so there is a widening gap between the proliferation of satellite imagery and the limited availability of high-quality labels \cite{seneviratne2021self, wang2022self}.
Another problem is that the target detection object is obscured.
Sidewalks are always adjacent to trees, so they are often partially or completely blocked in remote sensing images, i.e., canopy occlusion in the overhead images \cite{senlet2012satellite}.
In this case, we still want our pipeline to identify the part of the footpath that is obscured in order to transform the segmented footpath into a whole network.

To address these issues, this work introduces a pipeline for constructing a footpath network from remote sensing images of a given area only requiring a minimum of manually produced annotations.
As shown in Fig.~\ref{fig:intro}, the real sidewalk network map is the top image, while the sidewalk map automatically generated by our proposed method is at the bottom.
Considering that supervised training requires considerable labeled data, we use a self-supervised learning method to pre-train our model on an unlabeled remote sensing dataset, which is known to be an effective strategy for segmentation tasks\cite{thisanke2023semantic}.
Then, we obtain a generalizable feature extractor for the subsequent footpath segmentation task. 
We then attempt to solve for the problem of occluded segmented objects in terms of dataset construction.
When annotating the segmentation task, we ignore the occlusion and directly generate the corresponding real footpath mask for the image.
In conclusion, our contributions are: first, we create two datasets for pre-training and fine-tuning separately, then we present our pipeline which is able to generate footpath maps with minimal manual effort (Sec.~\ref{sec3}). 
We evaluate our segmentation model on both validation and test set and our model obtains better F1-score and mIoU results than the supervised pre-training baseline model.
Our quantitative and visualized results on a remote sensing dataset in the Melbourne area are present in Sec.~\ref{sec4}.
In particular, our pipeline can be easily extended to new datasets or other map-building tasks at a very low cost.

\section{Related Work}
\subsection{Self-supervised Learning}
The main goal of self-supervised learning is to learn task-independent representations of the input data that can be easily generalized to downstream tasks \cite{wang2022self}.
The self-supervised model designs a pretext task whose annotations can be obtained from the dataset automatically without the need for manual annotations or a labeled dataset.
Contrastive learning, a type of self-supervised method, has recently attracted a lot of attention in the field of self-supervised learning and become an essential component for natural language process, computer vision, and other domains \cite{2020survey}.
This method does not rely on a single and specific pretext task, it aims to make similar images closer and different images far away from each other in the feature space.
Therefore, it avoids learning task specific representations and is able to generalize well in different downstream tasks such as classification, detection, segmentation, and so on.
\cite{grill2020bootstrap} proposes Bootstrap Your Own Latent (BYOL) consisting of two neural networks named online and target which take positive pairs as the input and learn from each other.
A simple but effective data augmentation named multi-crop is proposed by \cite{caron2020unsupervised} which uses a series of images of different resolutions instead of augmented images with a fixed resolution.
Inspired by BYOL, \cite{caron2021emerging} introduces a simple contrastive learning method with a form of self-distillation with no labels (DINO).
They observe that the features extracted by a self-supervised Vision Transformer (ViT) trained by DINO deliver useful knowledge for semantic segmentation.
Based on DINO and multi-crop, \cite{wanyan2023dino} proposes DINO-MC which uses multi-size local crops instead of a series of fixed-size local crops which is proven that the features extracted are more useful than DINO on some remote sensing tasks.
There has been some work on the utilization of computer vision methods based on remote sensing imagery to generate footpath or sidewalk maps, but most of them are based on supervised methods.
The application of the self-supervised model which is capable of achieving comparable results even requiring less labeled images still lacks exploration in this field.
This work applies DINO-MC \cite{wanyan2023dino}, a self-supervised learning method specialized for remote sensing imagery by considering the variation in feature sizes present in remote sensing imagery, to learn effective representations for footpath segmentation.

\subsection{Footpath Segmentation}
Physical site or aerial image surveys can generate accurate and high-quality footpath maps, but they are also a time-consuming and laborious work \cite{proulx2015database}.
With the development of remote sensing technology, many automatic mapping tools have been proposed, like pedestrian Global Positioning System (GPS) trajectories and airborne Light Detection and Ranging (LiDAR).
While GPS trajectories of pedestrians can result in lower costs of data collection, it has limited accuracy, in contrast, LiDAR has higher geometric accuracy but also higher cost \cite{hosseini2023mapping}.
Therefore, many existing works \cite{schlosser2016fusing, matti2017combining, alfred2023fully, hou2020network} have explored combining LiDAR technology with deep learning models.
However, these methods still require specialized equipment and are labor-intensive, and our goal is to propose a generic, low-threshold, practical, and scalable pipeline for certain map generation tasks.
In line with our work, a few papers focus on applying computer vision techniques only to extract footpath maps from remote sensing imagery \cite{luo2019developing, ning2022sidewalk, hosseini2023mapping}.
However, these strategies rely on extra different views of images or a large amount of labeled data support for training, which increase the threshold for other specific applications, like combining horizontal disparity, height above ground, and angle (HHA) features with RGB-D image features to get more accurate maps than using HHA or RGB only \cite{schlosser2016fusing}. In contrast, our work uses self-supervised learning and a much smaller dataset (1000 images or less) for training, thus considerably reducing labeling requirements.
Considering the convenience and accessibility of a large amount of remote sensing image data nowadays, it is worth exploring the use of computer vision methods to automatically generate footpath maps from the remote sensing images only, without the need for large amounts of annotated data.
While some work has attempted to explore this as a classification problem \cite{seneviratne2022urban}, we argue that it is important to predict segmentation results for generating fine-grained spatial insights and measurement.

\section{Methodology}
\label{sec3}
\begin{figure}
  \centering
  \includegraphics[width=0.7\linewidth]{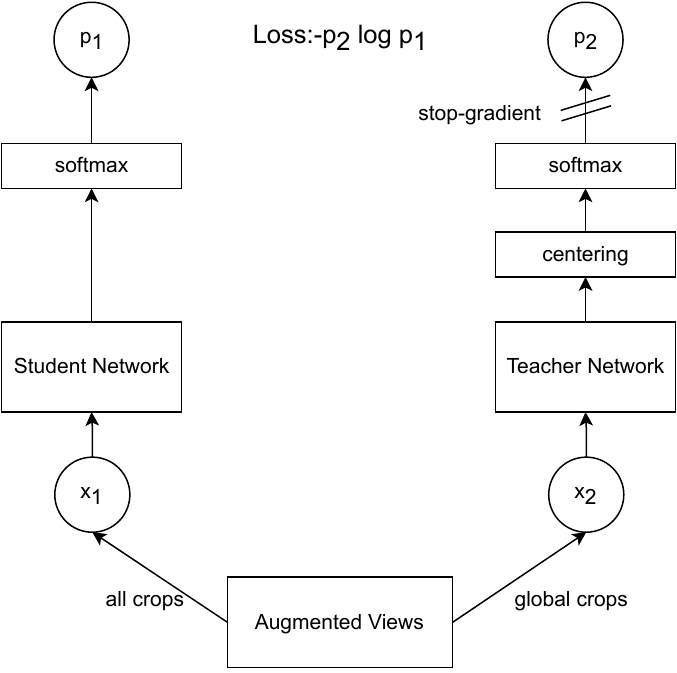}
  \caption{
  The structure of DINO-MC used in this work.
  A self-supervised contrastive model with knowledge distillation.
  }
  \label{fig:DINO-MC}
\end{figure}
\begin{figure*}
  \centering
  \includegraphics[width=0.9\linewidth]{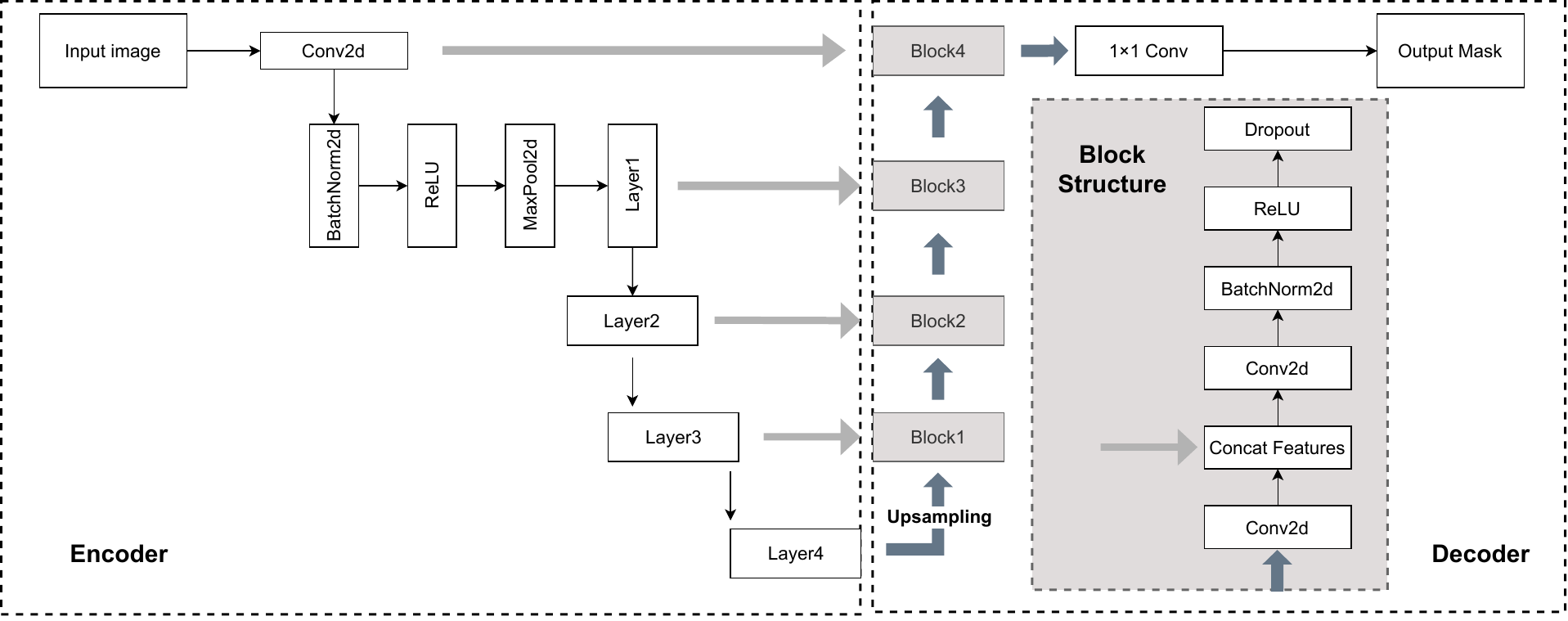}
  \caption{
  The structure of the U-Net used in this work consists of two modules.
  The left part of the figure is the encoder module of the U-Net.
  In this work, we use the pre-trained Wide ResNet as the encoder to calculate the representations of the inputs.
  The right part of the figure is the decoder module of the U-Net.
  It is a custom convolutional neural network used to reconstruct the mask.
  The decoder module is composed of four blocks, and the structure of each block is also shown in this figure. 
  The block takes both the output of the previous layer and the output of the encoder layer as the input and concatenates them together to process.
  }
  \label{fig:U-Net}
\end{figure*}
Our pipeline consists of two phases, image segmentation (mask generation) and GIS layer generation.
To generate masks, we first pre-train a convolutional neural network for feature extraction in a self-supervised manner.
Then we use an encoder-decoder model named U-Net to detect and segment the footpath mask for a specific area of remote sensing images.
In the second phase, we extract the polygons from the generated masks and stitch them into a whole network according to their latitude and longitude coordinates.

\subsection{Datasets}
In this work, we build two different remote sensing imagery datasets for the whole pipeline.
Both datasets (the self-supervised training set and the footpath segmentation set) are downloaded from MetroMap, a provider of high-resolution aerial imagery that is updated frequently.
Our datasets are collected based on XYZ Tiles, which is also known as slippy map tiles.
It is a system using Web Mercator coordinates, X and Y represent the index and Z represents the zoom level of the tiles.
The collected images are $256 \times 256$ pixel.
We request the slippy tiles with $Z=21$ from MetroMap.
The training set of the self-supervised representation learning consists of 100,000 unlabeled remote sensing images of Sydney, Australia.
We collect approximately 10 million remote sensing images of Melbourne in 2020 for the downstream task of sidewalk segmentation.
The footpath segmentation masks are generated automatically by converting the geographic network into binary masks.
We download the footpath geographic network of a small area of Melbourne and partition this network into a series of small polygons.
These polygons are segmented based on latitude and longitude coordinates which correspond to the downloaded remote sensing image tiles.
The segmented polygons are then converted to binary images and used as ground truth masks for remote sensing images with the same X and Y coordinates.
In this way, we generated footprint masks for a total of 40,363 images.
Among the labeled data, 1200 images with good annotations are manually selected, of which 1000 are used as the training set and 200 as the validation set.
The remaining 39,163 labeled images are used as the test set.

\subsection{Models}
\textbf{Self-supervised model.} 
The labeling of remote sensing image tasks, especially the segmentation task, is time-consuming, labor-intensive, and requires expertise.
Loading pre-trained model weights, rather than initializing with random weights, can accelerate model convergence and improve performance.
However, applying the traditional supervised method to pre-train the model requires a large amount of labeled data.
So in this work, we choose to use a self-supervised approach named DINO-MC to pre-train the model to learn a general feature representation for remote sensing images.
After pre-training, the model is able to have an initial ability for feature extraction, which significantly reduces the amount of training data required for the downstream task.
DINO-MC is a contrastive self-supervised method that does not depend on a specific pretext task.
Relying on a single pretext task can only learn pretext-specific features, which may lead to poor generalization of the model.
DINO-MC utilizes the knowledge distillation architecture instead of a single pretext task, to train the model to learn the relationship between the whole and the parts and capture the essential features that do not change when using various image augmentations.
As illustrated in Fig.~\ref{fig:DINO-MC}, there are two networks in DINO-MC named teacher and student networks which have identical architecture but different weights.
The inputs of the student network include different sizes of crops (global and local crops) of the initial image, while the teacher network is only fed with two global crops of the same size.
In addition to the different sizes, the input crops are applied with different and random data augmentation methods, including Gaussian Blur, color jitter, solarization, and flip.
The goal of the training process is to make the image representations extracted by two networks as similar as possible in the feature space.
Therefore, the model does not require any labeled datasets to learn a generic representation that is invariant to different augmentations.
Compared to DINO, DINO-MC employs different sizes of local crops instead of a single size which is proven to improve the performance of the representations learned by the model on different downstream tasks \cite{wanyan2023dino}.

Since Wide ResNet was proven to achieve better results than ResNet from the results of \cite{wanyan2023dino}, in this task, we apply Wide ResNet as the backbone (teacher and student network) of DINO-MC. 

\textbf{Segmentation model.}
We choose U-Net, a U-shape segmentation network, to do the footpath segmentation task.
It takes an image as input, outputs a dense prediction that assigns a category to each pixel, i.e., a binary mask showing where the footpath is.
It mainly consists of two modules named encoder and decoder.
In addition, a special structure of U-Net is the skip connection that connects the shallow features to the decoder directly.
The pre-trained backbone model Wide ResNet is used as the encoder to generate the representations for the input images.
The output of the first convolution layer and the last four blocks of Wide ResNet are saved and passed to the corresponding layers of the decoder module which is called skip-connection, so the semantics information can be forwarded to deep layers.
The decoder used in U-Net is a custom convolutional neural network for generating the segmentation masks by upsampling the feature maps.

\textbf{Implementation details. }
In this section, we present more details about the implementation in the experiments.
For self-supervised training, we pre-train the self-supervised model on $100,000$ unlabelled remote sensing images.
The optimizer we use to update the weights of the model is adamw.
During the training process, the batch size is set to $16$ per GPU, and four GPUs are employed in total.
Following DINO, we apply the learning rate warmup for the first $10$ epochs, during which the learning rate will linearly increase. 
Then the learning rate starts to decrease following a cosine schedule.
Following DINO-MC, we crop the input image into two global crops of the same size $224 \times 224$ and six local crops of different sizes which are $184\times184, 164\times164, 144\times144, 124\times124, 104\times104$, which is called multi-crop.  
We use the bicubic interpolation to resize images which is the same crop setting as DINO.
After cropping, we apply HorizontalFlip, color jittering and GaussianBlur on the generated crops then additionally apply the Solarization on one of the global crops.

In the segmentation task fine-tuning, the implementation of the U-Net refers to the codes of SeCo \cite{manas2021seasonal}, which is mainly based on the Pytorch Lightning, a deep learning framework.
We initialize the Wide ResNet with the pre-trained weights and apply it as the encoder of the U-Net to do feature extraction.
During fine-tuning, the batch size is $32$ on a single GPU and the learning rate is $6e-5$.
The fine-tuning process may result in feature loss, and we need to retain as many useful features learned from pre-training as possible, so the learning rate cannot be set too large.
The loss function proved to be the best model in our experiments is dice\_bce\_loss \cite{zhou2018dlinknet}, which combines the BCE (binary cross entropy) and dice coefficient.
The parameters of the segmentation model are updated by two different experimental schemes.
The first one is to freeze the feature extractor (Wide ResNet) to compute the representations of the input images and only adapt the weights of the decoder network.
Another is to update the whole U-Net including both the pre-trained encoder and the custom decoder networks.

\subsection{Evaluation}

We perform experiments with our model by applying both end-to-end fine-tuning and the encoder-frozen fine-tuning.
The main object of the assessment is the footpath segmentation results.
Because of the imbalance in the number of categories (at the pixel level), the pixel accuracy is not able to accurately reflect the performance of the segmentation task.
Therefore, we employ both F1-score and mean Intersection over Union (mIoU) evaluation metrics to calculate how well the generated masks match the ground truth masks.

F1-score is the harmonic mean of precision and recall.
The calculation of F1-score is shown in Eq.\ref{eq3.1}, \ref{eq3.2}, and \ref{eq3.3}, with $TP$ denoting the number of positive examples that are properly predicted, $FP$ denoting the number of positive instances that are wrongly forecasted, and $FN$ denoting the number of negative cases that are incorrectly predicted.
mIoU, also referred to as the Jaccard Index, is one of the most widely used assessment measures for segmentation tasks.
As shown in Eq.\ref{eq3.4}, IoU is the number of pixels that overlap between the generated segmentation mask and the ground truth mask divided by the number of their union pixels.

\begin{equation}
Precision = \frac{TP}{TP+FP}
\label{eq3.1}
\end{equation}

\begin{equation}
Recall = \frac{TP}{TP+FN}
\label{eq3.2}
\end{equation}

\begin{equation}
F1-score = 2 \times \frac{Precision \times Recall}{Precision + Recall}
\label{eq3.3}
\end{equation}

\begin{equation}
IoU = \frac{overlap\_pixels}{union\_pixels}
\label{eq3.4}
\end{equation}


\subsection{GIS Layer Generation}
After the image segmentation phase is completed, we obtain the predicted footpath masks in raster format.
As shown in Fig.~\ref{fig:pipeline}, we first extract the footpath contour from raster images and get the coordinates of the contour.
We calculate the coordinates of the contours in the image and translate them into real-world latitude and longitude.
The contour represented by real coordinates can then be converted into a polygon.
The generated polygons are the geometry objects and can be combined and processed according to their locations.
We apply the simplify method
Finally, the  saved in GeoJSON files or shapefiles by an open-source Python tool named GeoPandas.
The produced file is able to be loaded and operated as a layer of the subsequent project across multiple GIS software applications.

\section{Experiments and Results}
\label{sec4}
The experiments mainly focus on the footpath segmentation to generate the binary mask for the input remote sensing imagery.
We utilize two fine-tuning methods to train our model on different sizes of training sets and the fine-tuned models are evaluated on the validation set quantitatively, and there is no overlap between the training and validation sets.
Two baseline models are applied to the footpath segmentation task, and we only fine-tune them on 1000 training images to compare with our model.
Then we provide some visualization results of this work.

\begin{table}
\caption{
  Applying two fine-tuning strategies on different sizes of datasets.
  We build four footpath segmentation training sets of different sizes (containing 100, 400, 500, and 1000 images respectively) to explore the relationship between the model performance and the size of the training set.
  The validation set consists of 200 images with no overlap with the training set. 
  }
  \centering
  \begin{tabular}{lccc}
    \hline
    \#images & Decoder (val) & All (val) & All (test) \\
    \hline
    100 & 57.00 & 63.11 & 51.06 \\
    400 & 62.02 & 71.84 & 59.03 \\
    500 & 64.33 & 72.14 & 60.25 \\
    1000 & 68.35 & 76.58 & 63.97 \\
    \hline
  \end{tabular}
  \label{tab:finetune1}
\end{table}

\begin{figure}
  \centering
  \includegraphics[width=0.7\linewidth]{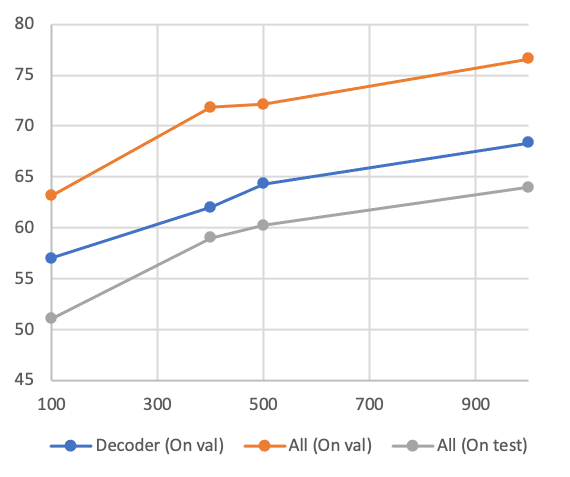}
  \caption{
  F1-scores for models fine-tuned on training sets of different sizes.
  }
  \label{fig:size_finetune}
\end{figure}

\subsection{Quantitative Results}

\textbf{Comparing models fine-tuned on different-sized training sets.}
A machine learning model's performance is thought to be significantly influenced by the dataset size \cite{althnian2021impact}.
We create datasets randomly in four different sizes including 100, 400, 500, and 1000, and the smaller datasets are incorporated into the larger datasets.
We load the Wide ResNet pre-trained in DINO-MC as the encoder of U-Net and experimented with two fine-tuning modes, one is to freeze the encoder and update only the decoder module, and the other is to update all parameters of U-Net.
The quantitative results are shown in Tab.~\ref{tab:finetune1}.
Decoder (val) is updating decoder only and evaluated on the validation set after fine-tuning.
All (val) represents updating all parameters of U-Net and evaluated on the validation set, and all (test) is the same model but evaluated on the test set. 
The results shown in this table are the F1-score of different models evaluated on the same validation set (containing a total of 200 images) and the same test set (consisting of 39,163 images in total).
We fine-tune our model on four sizes of training sets and list the F1-score on validation and test sets.
Comparing the F1-score of the two fine-tuning strategies on the validation set, updating the parameters of the entire network achieves better results in the quantitative evaluation.

When fine-tuning on 100, 400, 500, and 1000 images respectively, updating all parameters achieves higher F1-scores of $6.11, 9.82, 7.81$, and $8.23$ than only updating the decoder model.
From Fig.~\ref{fig:size_finetune}, the F1-scores achieved by the model become progressively larger as the training data increases.
When the number of images increases from 100 to 500, the improvement of the model is obvious: the F1-scores of updating all parameters increase by $9.03$ on the validation set and $9.19$ on the test set, the F1-score of updating the decoder parameters only increase by $7.33$ on the validation set.
But when the number of images increases from 500 to 1000, the F1-score of updating all parameters increase by $4.44$ on the validation set and $3.72$ on the test set, the F1-score of updating the decoder parameters only increase by $4.02$ on the validation set.
Therefore, the performance of both models improves significantly when the number of images increases from 100 to 500, but the improvement reduces when the number of images increases from 500 to 1000.

\textbf{Comparing with baseline models.}
\begin{table}
\caption{Comparison between baselines and our model.}
  \centering
  \begin{tabular}{lccc}
    \hline
    Model & Dataset & F1-score & mIoU \\
    \hline
    Random WRN101 & Val & 63.19 & 46.4 \\
    Random WRN101 & Test & 50.11 & 33.49 \\
    ImageNet1K WRN101 & Val & 69.77 & 53.7 \\
    ImageNet1K WRN101 & Test & 55.86 & 38.81 \\
    DINO-MC WRN101 & Val & \textbf{76.58} & \textbf{62.2} \\
    DINO-MC WRN101 & Test & \textbf{63.97} & \textbf{47.09} \\
    \hline
  \end{tabular}
  \label{tab:finetune2}
\end{table}
We utilize Wide ResNet initialized with different pre-trained weights as the encoder of U-Net to extract features of the input image.
Two baseline models are applied in this experiment.
One is Wide ResNet without any pre-training which is loaded with random weights. 
The other is pre-trained on ImageNet1K in the supervised manner. 
Tab.~\ref{tab:finetune2} provides the quantitative results of two baseline models and our best performance model.
These three models are fine-tuned and evaluated on the same footpath segmentation dataset with 1000 training images, 200 validation images, and 39,163 test images.
The first two models listed in the table are the baseline models.
Random WRN101 is the Wide ResNet initialized with random weights, while ImageNet1K WRN101 is the Wide ResNet pre-trained on ImageNet1K in a supervised manner.
DINO-MC WRN101 is the Wide ResNet pre-trained as the backbone of DINO-MC in a self-supervised manner.

From the table, Wide ResNet pre-trained in DINO-MC achieves better results of F1-score and mIoU metrics on both validation and test sets than other two baseline models.
DINO-MC WRN101 achieves
$6.81$ and $13.39$ higher F1-score than ImageNet1k WRN101 and Random WRN101 on validation set, and $8.11$ and $13.86$ higher F1-score on test set.
DINO-MC WRN101 achieves
$8.5$ and $15.8$ higher mIoU than ImageNet1k WRN101 and Random WRN101 on validation set, and $8.28$ and $13.6$ higher mIoU on test set.

\begin{figure}
     \centering
     \subfigure{
        \centering
        \includegraphics[width=0.7\linewidth]{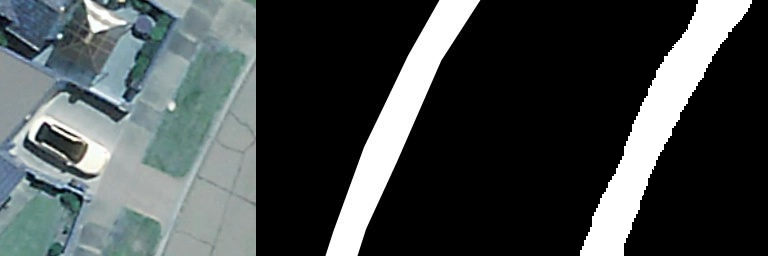}
     }
     \subfigure{
         \centering
         \includegraphics[width=0.7\linewidth]{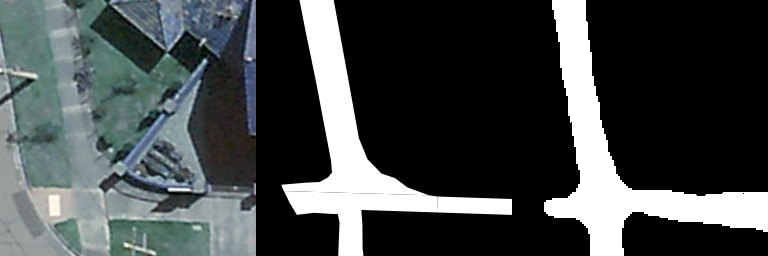}
     }
     \subfigure{
         \centering
         \includegraphics[width=0.7\linewidth]{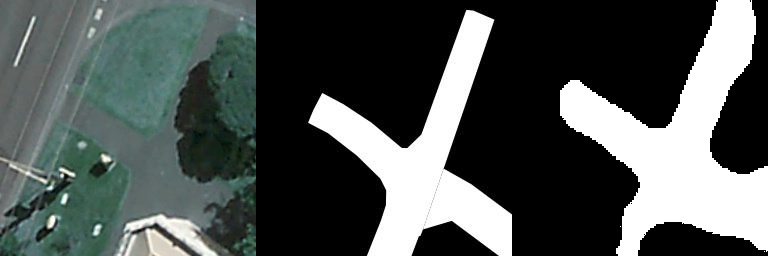}
     }
     \subfigure{
         \centering
         \includegraphics[width=0.7\linewidth]{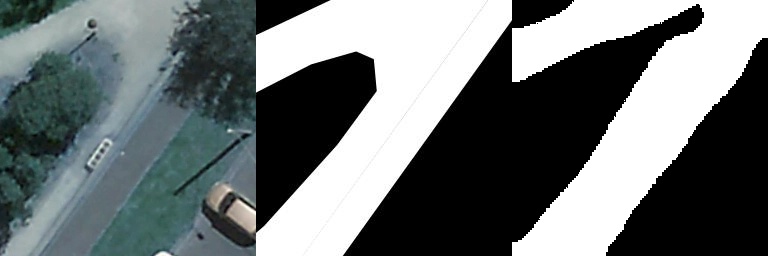}
     }
     \subfigure{
         \centering
         \includegraphics[width=0.7\linewidth]{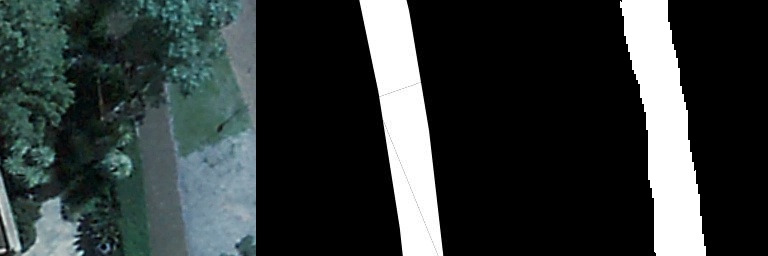}
     }
     \caption{
    Examples of the mask output from the best performing fine-tuning model.
    The model is pre-trained as the backbone of DINO-MC in a self-supervised manner and then fine-tuned on footpath segmentation task training set with 1000 remote sensing images.
    The first column (from left to right) shows the original remote sensing images, the second row shows the ground truth masks, and the third row shows the generated binary masks. }
  \label{fig:visualization_instances}
\end{figure}

\begin{figure}
  \centering
  \includegraphics[width=0.9\linewidth]{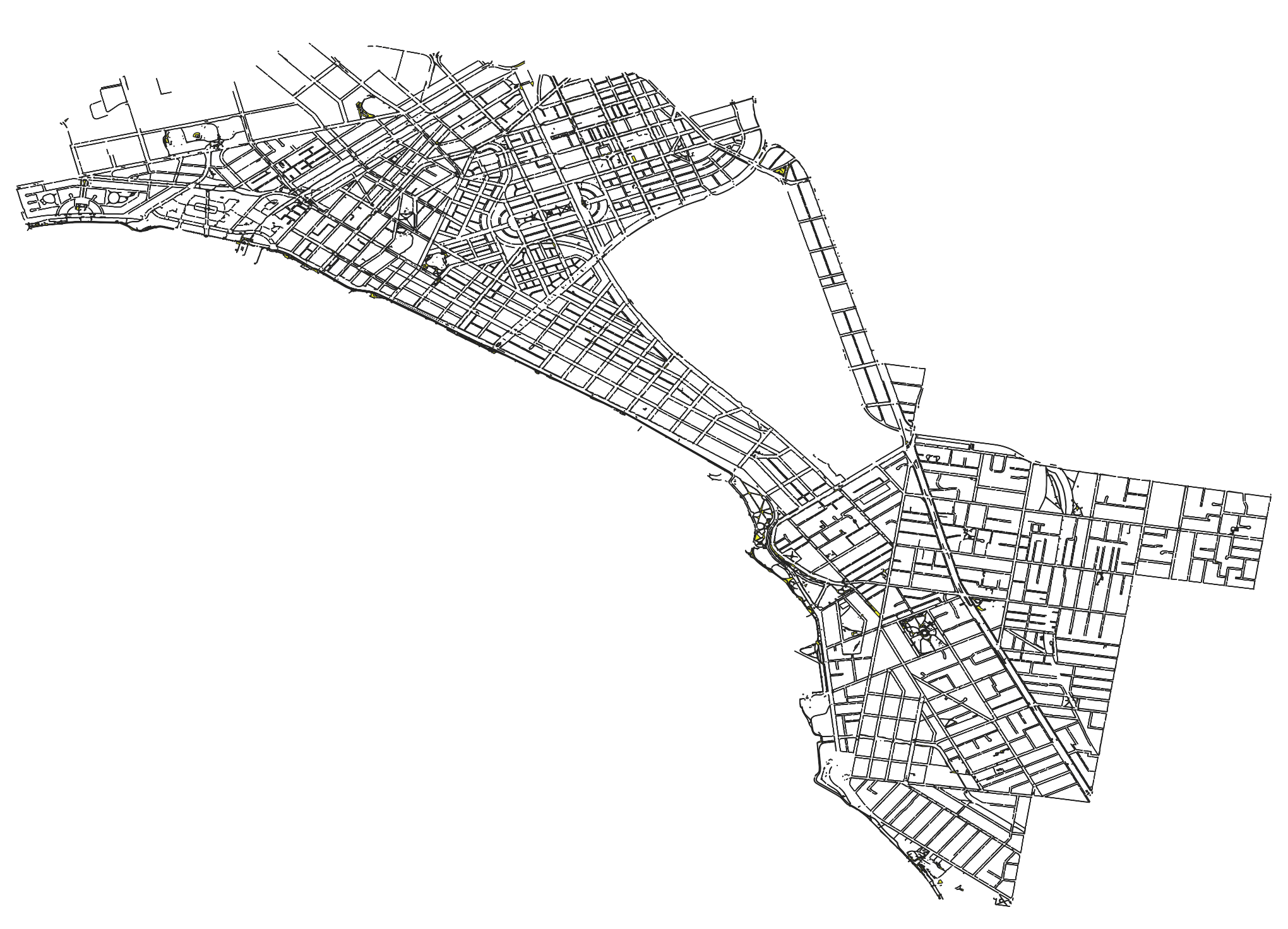}
  \includegraphics[width=0.9\linewidth]{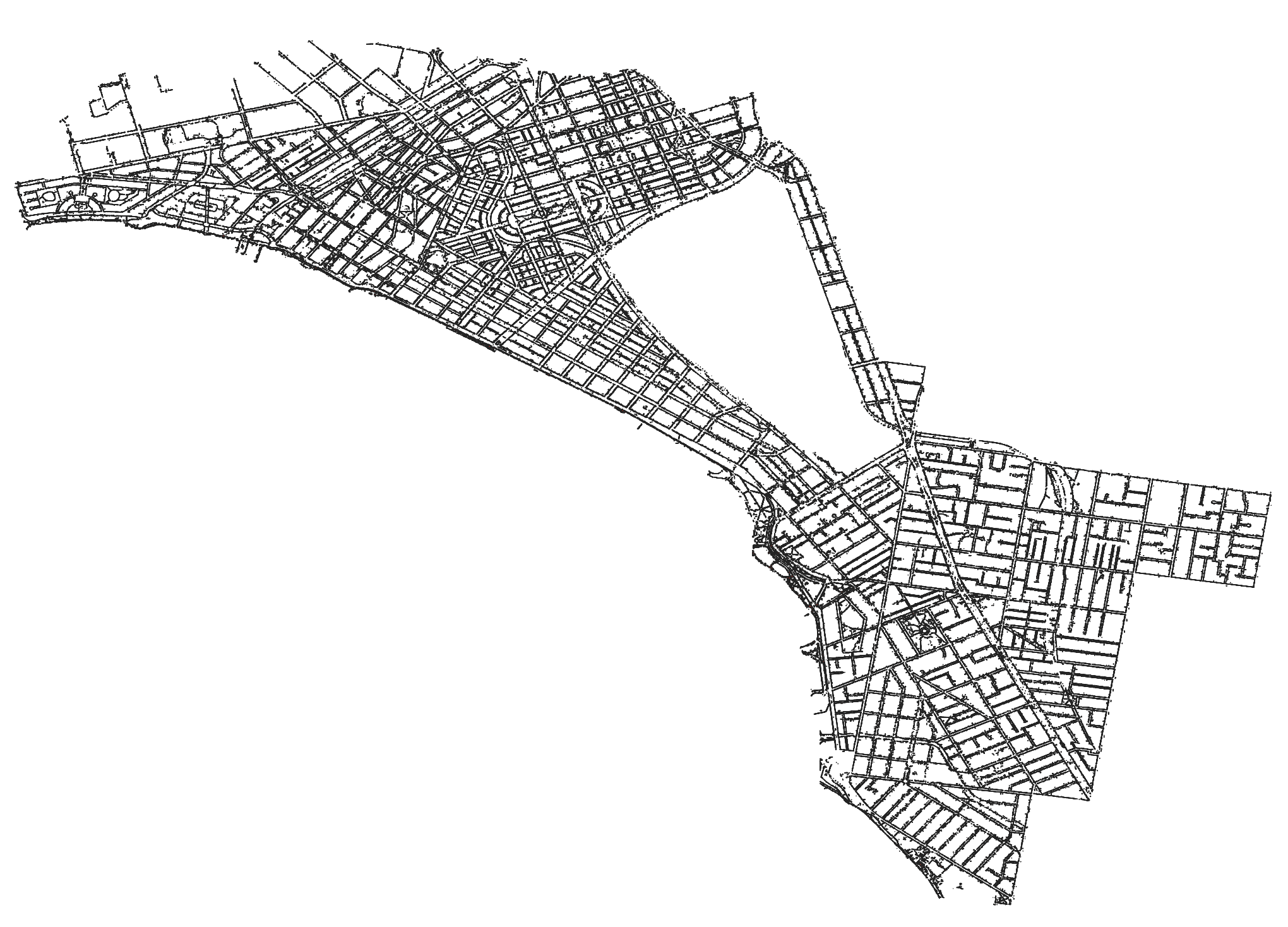}
  \caption{
  Visualization of the whole generated network.
  Top: the ground truth footpath network in an area of Melbourne.
  Bottom: the generated footpath network using our pipeline.
  }
  \label{fig:visualize_network}
\end{figure}
\subsection{Visualization}
We use the fine-tuned U-Net with Wide ResNet pre-trained in DINO-MC for mask generation which achieves the best quantitative results on both validation and test sets.
Fig.~\ref{fig:visualization_instances} shows five instances of the footpath segmentation task, where the first column is the input remote sensing image of the model, the second column is the ground truth mask, and the last column presents the output mask of our pipeline.
From the visualization results of the generated masks, we can find that our model is capable of detecting the footpath from the remote sensing imagery and restoring occluded and missing sidewalks even in the presence of tree canopy occlusion (see the third row of instances in Fig.~\ref{fig:visualization_instances}).
Our model can detect the location and shape of the footpath from remote sensing images, but there are prediction errors in the specific width as well as in the edges.

Fig.~\ref{fig:visualize_network} presents the entire ground truth footpath map (on the top) and the entire generated footpath map (on the bottom).
The generated map is able to depict the whole network structure of this area, which is quite similar to the ground truth.

\subsection{Discussion and Error Analysis}
From the quantitative results, we find that pre-training on larger datasets can produce better results, but this improvement reduces as the amount of data increases.
Updating all the parameters during fine-tuning phase achieves better results than freezing the encoder model, since during this process, the features extracted by the encoder generalize to the specific footpath downstream task which helps the encoder construct better segmentation mask.

We observe that when updating all parameters of U-Net, the improvement resulting from increasing the size of the training set is greater than when updating only the decoder module.
One possible reason is that when freezing the encoder module, the fine-tuned model contains only the decoder, which can be seen as training a smaller model than training the whole U-Net, and the feature learning ability of the smaller model is to some extent more limited compared to the larger model.
Compared to the random initialized Wide ResNet, our self-supervised Wide ResNet gains better performance which proves the effectiveness of the pre-training process again.
We even outperform the supervised baseline model showing the large potential of applying self-supervised learning in GIS map generation based on remote sensing imagery.

From the visualization results, our pipeline is able to identify the shape and location of the footpath, and the errors are mainly in the prediction of the specific sidewalk's precise width prediction and the edge segmentation.
The possible reason could be that the zoom level of the remote sensing imagery tiles we use for segmentation input is large, and each image only covers a small area.
Therefore, for each image, the segmentation model can observe and learn less contextual information.
Another possible reason is the problem of occluded segmented objects.
Due to the camera angle, lighting problems, and the diversity of occlusions, the model suffers more interference in dealing with the contour of the footpath.

In the future - by combing our outputs with existing data collection processes such as OSM, and Captial Works programs (within Councils) - the output quality will be progressively improved enabling increasingly robust urban analytic outputs and improving data drived urban design decision-making.
\section{Conclusion}
This paper has shown the potential of applying a self-supervised model to footpath map generation only using remote sensing imagery. 
We propose a pipeline for generating a geographic footpath map only based on the corresponding remote sensing images.
First, we employ a self-supervised learning model DINO-MC to train the Wide ResNet to learn general feature representations.
Then we load the pre-trained Wide ResNet as the encoder of the segmentation model U-Net and fine-tune it on the footpath segmentation task.
After training, the best-performing fine-tuned model is applied to the remote sensing imagery of a specific region to obtain the generated masks of raster format.
Based on the masks, we extract the contour of the predicted footpath and convert them to polygons, which are saved in the GeoJSON files for the following application.
Our approach is highly automated, has a low threshold, and is ready to extend to other datasets or applications.
Our model achieves better F1-score and mIoU than the supervised WideResNet baseline model. 

\section*{Acknowledgment}
This research was supported by The University of Melbourne’s Research Computing Services and the Petascale Campus Initiative. 
This research was supported (partially or fully) by the Australian Government through the Australian Research Council's Centre of Excellence for Children and Families over the Life Course (Project ID CE200100025).

{\small
\bibliographystyle{IEEEtran}
\bibliography{footpath_bib}
}
\end{document}